\documentclass[10pt,twocolumn,letterpaper]{article}

\usepackage{cvpr}
\usepackage{times}
\usepackage{epsfig}
\usepackage{graphicx}
\usepackage{amsmath}
\usepackage{amssymb}
\usepackage{subfiles}
\usepackage{multirow}
\usepackage{booktabs}

\usepackage[pagebackref=true,breaklinks=true,letterpaper=true,colorlinks,bookmarks=false]{hyperref}

\cvprfinalcopy 

\def\cvprPaperID{4322} 

\ifcvprfinal\fi
\begin{document}

\title{Vision-Language Navigation with Self-Supervised Auxiliary Reasoning Tasks}

\author{Fengda Zhu$^1$\hspace{6mm}
Yi Zhu$^2$\hspace{6mm}
Xiaojun Chang$^1$ \hspace{6mm}
Xiaodan Liang$^{3,4}$ \\
%
$^1$Monash University \hspace{2mm} $^2$University of Chinese Academy of Sciences \\
$^3$Sun Yat-sen University \hspace{2mm} $^4$Dark Matter AI Inc.  \\
{\tt\small zhufengda@yahoo.com \hspace{2mm} zhu.yee@outlook.com } \\
{\tt\small cxj273@gmail.com \hspace{2mm} xdliang328@gmail.com }
}

\maketitle
\thispagestyle{empty}

\begin{abstract}
Vision-Language Navigation (VLN) is a task where agents learn to navigate following natural language instructions. The key to this task is to perceive both the visual scene and natural language sequentially. Conventional approaches exploit the vision and language features in cross-modal grounding. However, the VLN task remains challenging, since previous works have neglected the rich semantic information contained in the environment (such as implicit navigation graphs or sub-trajectory semantics). In this paper, we introduce Auxiliary Reasoning Navigation (AuxRN), a framework with four self-supervised auxiliary reasoning tasks to take advantage of the additional training signals derived from the semantic information. The auxiliary tasks have four reasoning objectives: explaining the previous actions, estimating the navigation progress, predicting the next orientation, and evaluating the trajectory consistency. As a result, these additional training signals help the agent to acquire knowledge of semantic representations in order to reason about its activity and build a thorough perception of the environment. Our experiments indicate that auxiliary reasoning tasks improve both the performance of the main task and the model generalizability by a large margin. Empirically, we demonstrate that an agent trained with self-supervised auxiliary reasoning tasks substantially outperforms the previous state-of-the-art method, being the best existing approach on the standard benchmark\footnote[1]{VLN leaderboard: https://evalai.cloudcv.org/web/challenges/ \\ 
	challenge-page/97/leaderboard/270}. 
\end{abstract}
   %
   %

\section{Introduction}
 Increasing interest rises in Vision-Language Navigation (VLN)~\cite{anderson2018vision} tasks, where an agent navigates in 3D indoor environments following a natural language instruction, such as \textit{Walk between the columns and make a sharp turn right. Walk down the steps and stop on the landing}. 
The agent begins at a random point and goes toward a goal by means of active exploration. A vision image is given at each step and a global step-by-step instruction is provided at the beginning of the trajectory. 

Recent research in feature extraction~\cite{he2015deep, anderson2018bottom, mikolov2013distributed, ren2015faster, zeng2020dense}, attention~\cite{anderson2018bottom, devlin2018bert,lu2016hierarchical} and multi-modal grounding~\cite{antol2015vqa, lu2019vilbert, tan2019lxmert} have helped the agent to understand the environment. Previous works in Vision-Language Navigation have focused on improving the ability of perceiving the vision and language inputs~\cite{gupta2017cognitive, fried2018speaker, wang2018look} and cross-modal matching~\cite{wang2018reinforced, zhu2020vision}. With these approaches, an agent is able to perceive the vision-language inputs and encode historical information for navigation. 

\begin{figure}[t]
	\centering
	\includegraphics[width=0.95\linewidth]{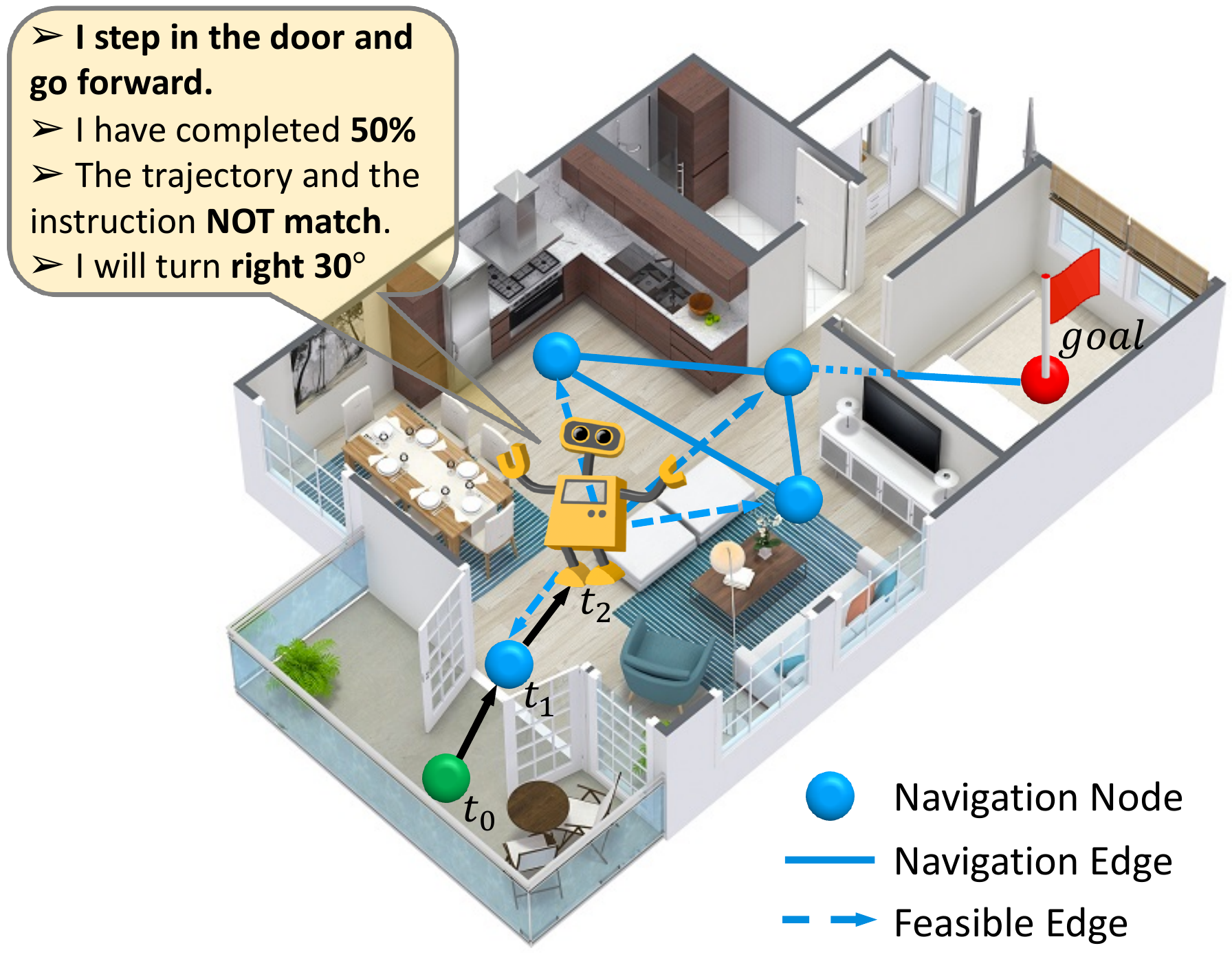}
	\caption{
	A simple demonstration of an agent learning to navigate with auxiliary reasoning tasks. 
	The green circle is the start position and the red circle is the goal. Four nodes are reachable by the agent in the navigation graph. Auxiliary reasoning tasks (in the yellow box) help the agent to infer its current status. }
	\label{fig:demo}
	\vspace{-0.5cm}
\end{figure}


However, the VLN task remains challenging since rich semantic information contained in the environments is neglected: 
1) Past actions affect the actions to be taken in the future. To make a correct action requires the agent to have a thorough understanding of its activity in the past. 
2) The agent is not able to explicitly align the trajectory with the instruction. Thus, it is uncertain whether the vision-language encoding can fully represent the current status of the agent. 
3) The agent is not able to accurately assess the progress it has made. Even though Ma \emph{et al.}~\cite{ma2019self} proposed a progress monitor to estimate the normalized distance toward the goal, labels in this method are biased and noisy. 
4) The action space of the agent is implicitly limited since only neighbour nodes in the navigation graph are reachable. Therefore, if the agent gains knowledge of the navigation map and understands the consequence of its next action, the navigation process will be more accurate and efficient. 

We introduce auxiliary reasoning tasks to solve these problems. There are three key advantages to this solution. 
First of all, auxiliary tasks produce additional training signals, which improves the data efficiency in training and makes the model more robust. 
Secondly, using reasoning tasks to determine the actions makes the actions easier to explain. 
It is easier to interpret the policy of an agent if we understand why the agent takes a particular action. An explainable mechanism benefits human understanding of how the agent works. 
Thirdly, the auxiliary tasks have been proven to help reduce the domain gap between seen and unseen environments. It has been demonstrated~\cite{sun2019unsupervised, sun2019test} that self-supervised auxiliary tasks facilitate domain adaptation. Besides, it has been proven that finetuning the agent in an unseen environment effectively reducing the domain gap~\cite{wang2018reinforced, tan2019learning}. We use auxiliary tasks to align the representations in the unseen domain alongside those in the seen domain during finetuning.

In this paper, we introduce Auxiliary Reasoning Navigation (AuxRN), a framework facilitates  navigation learning. AuxRN consists of four auxiliary reasoning tasks: 
1) A \textbf{trajectory retelling task}, which makes the agent explain its previous actions via natural language generation;  
2) A \textbf{progress estimation task}, to evaluate the percentage of the trajectory that the model has completed;
3) An \textbf{angle prediction task}, to predict the angle by which the agent will turn next. 
4) A \textbf{cross-modal matching task} which allows the agent to align the vision and language encoding. 
Unlike ``proxy tasks"~\cite{lu2019vilbert, tan2019lxmert, su2019vl} which only consider the cross-modal alignment at one time, our tasks handle the temporal context from history in addition to the input of a single step. 
The knowledge learning of these four tasks are presumably reciprocal. 
As shown in Fig.~\ref{fig:demo}, the agent learns to reason about the previous actions and predict future information with the help of auxiliary reasoning tasks. 

Our experiment demonstrates that AuxRN dramatically improves the navigation performance on both seen and unseen environments. Each of the auxiliary tasks exploits useful reasoning knowledge respectively to indicate how an agent understands an environment. 
We adopt Success weighted by Path Length~(SPL)~\cite{anderson2018on} as the primary metric for evaluating our model. 
AuxRN pretrained in seen environments with our auxiliary reasoning tasks outperforms our baseline~\cite{tan2019learning} by 3.45\% on validation set. Our final model, finetuned on unseen environments with auxiliary reasoning tasks obtains 65\%, 4\% higher than the previous state-of-the-art result, thereby becoming the first-ranked result in the VLN Challenge in terms of SPL.

\section{Related Work}

\noindent \textbf {Vision-Language Reasoning} 
Bridging vision and language is attracting attention from both the computer vision and the natural language processing communities. 
Various associated tasks have been proposed, including Visual Question Answering (VQA)~\cite{agrawal2015vqa}, Visual Dialog Answering~\cite{thomason2019vision}, Vision-Language Navigation (VLN)~\cite{anderson2018vision} and Visual Commonsense Reasoning (VCR)~\cite{zellers2018from}. 
Vision-Language Reasoning~\cite{pearl1988probabilistic} plays an important role in solving these problems. 
Anderson \emph{et al.}~\cite{anderson2018bottom} apply an attention mechanism on detection results to reason visual entities. 
More recent works, such as LXMERT~\cite{tan2019lxmert}, ViLBERT~\cite{lu2019vilbert}, and B2T2~\cite{alberti2019fusion} obtain high-level semantics by pretraining a model on a large-scale dataset with vision-language reasoning tasks. 

\noindent\textbf{Learning with Auxiliary Tasks}
Self-supervised auxiliary tasks have been widely applied in the field of machine learning. 
Moreover, the concept of learning from auxiliary tasks to improve data efficiency and robustness~\cite{jaderberg2017reinforcement, pathak2017curiosity, veeriah2019discovery, ma2019self} has been extensively investigated in reinforcement learning. 
Mirowski \emph{et al.}~\cite{mirowski2017learning} propose a robot which obtains additional training signals by recovering a depth image with colored image input and predicting whether or not it reaches a new point. 
Furthermore, self-supervised auxiliary tasks have been widely applied in the fields of computer vision~\cite{zhao2017pyramid, gu2019scene, odena2017conditional}, natural language processing~\cite{devlin2018bert, lan2019albert} and meta learning~\cite{veeriah2019the, liu2019self}. Gidaris \emph{et al.}~\cite{gidaris2018unsupervised} unsupervisedly learn image features with a 2D rotate auxiliary loss, while Sun \emph{et al.}~\cite{sun2019test} indicate that self-supervised auxiliary tasks are effective in reducing domain shift. 

\noindent\textbf{Vision Language Navigation}
A number of simulated 3D environments have been proposed to study navigation, such as
Doom~\cite{kempka2016vizdoom}, AI2-THOR~\cite{kolve2017ai2} and House3D~\cite{wu2018building}.
However, the lack of photorealism and natural language instruction limits the application of these environments. 
Anderson \emph{et al.}~\cite{anderson2018vision} propose Room-to-Room (R2R) dataset, the first Vision-Language Navigation (VLN) benchmark based on real imagery~\cite{chang2017matterport3d}. 

\begin{figure*}[t]
	\centering
	\includegraphics[width=0.99\linewidth]{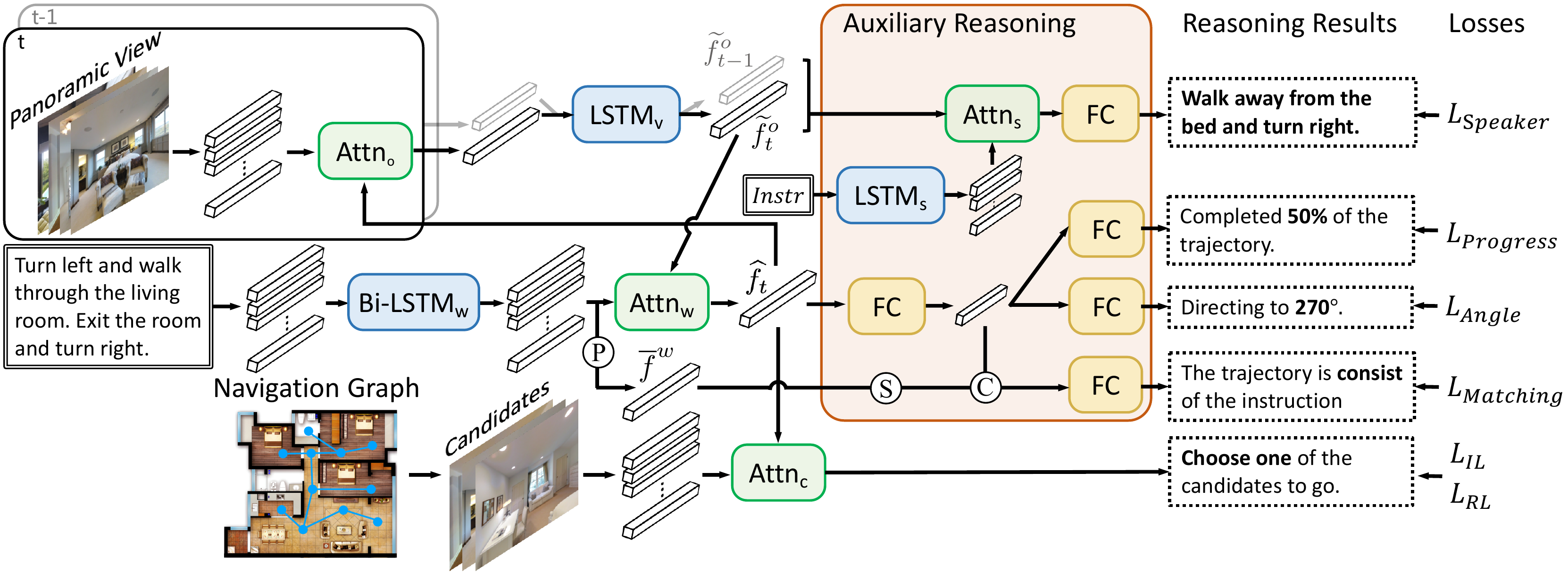}
	\caption{
	An overview of AuxRN. 
	The agent embeds vision and language features respectively and performs co-attention between them. The embedded features are given to reasoning modules and supervised by auxiliary losses. The feature produced by vision-language attention is fused with the candidate features to predict a action. 
	The ``P", ``S", and ``C" in the white circles stand for the mean pooling, random shuffle and concatenate operations respectively. }
	\label{fig:pipeline}
	\vspace{-0.5cm}
\end{figure*}

The Vision-Language Navigation task has attracted widespread attention since it is both widely applicable and challenging. Earlier work~\cite{wang2018look} combined model-free~\cite{mnih2016asynchronous} and model-based~\cite{racanire2017imagination}  reinforcement learning to solve VLN. Fried \emph{et al.} propose a speaker-follower framework for data augmentation and reasoning in supervised learning. In addition, a concept named ``panoramic action space" is proposed to facilitate optimization. Later work~\cite{wang2018reinforced} has found it is beneficial to combine imitation learning~\cite{bojarski2016end, ho2016generative} and reinforcement learning~\cite{mnih2016asynchronous, schulman2017proximal}. The self-monitoring method~\cite{ma2019self} is proposed to estimate progress made towards the goal. Researchers have identified the existence of the domain gap between training and testing data. Unsupervised pre-exploration~\cite{wang2018reinforced} and Environmental dropout~\cite{tan2019learning} are proposed to improve the ability of generalization. 

\section{Method}


\subsection{Problem Setup}\label{Problem Setup}

The Vision-and-Language Navigation (VLN) task gives a global natural sentence $I=\{w_0 ,..., w_l\}$ as an instruction, where each $w_i$ is a token while the $l$ is the length of the sentence. The instruction consists of step-by-step guidance toward the goal. At step $t$, the agent observes a panoramic view $O_{t}=\{o_{t,i}\}_{i=1}^{36}$ as the vision input. The panoramic view is divided into 36 RGB image views, while each of these views consists of image feature $v_i$ and an orientation description ($sin\ \theta_{t,i}$, $cos\ \theta_{t,i}$, $sin\ \phi_{t,i}$, $cos\ \phi_{t,i}$). For each step, the agent chooses a direction to navigate over all candidates in the panoramic action space~\cite{fried2018speaker}. Candidates in the panoramic action space consist of $k$ neighbours of the current node in the navigation graph and a stop action. Candidates for the current step are defined as $\{c_{t,1},...,c_{t,k+1}\}$, where $c_{t,k+1}$ stands for the stop action. Note that for each step, the number of neighbours $k$ is not fixed. 

\subsection{Vision-Language Forward}\label{Vision-Language Forward}
We first define the attention module, which is widely applied in our pipeline. 
Then we illustrate vision embedding and vision-language embedding mechanisms. At last, we demonstrate the approach of action prediction. 

\noindent\textbf{Attention Module}
At first we define the attention module, an important part of our pipeline. Suppose we have a sequence of feature vectors noted as $\{f_0 , ..., f_n\}$ to fuse and a query vector $q$. We implement an attention layer $\widehat{f} = \textnormal{Attn}(\{f_0 , ..., f_n\}  , q)$ as: 
\begin{equation}
    \begin{split}
    \alpha_i &= \textnormal{softmax}(f_i W_{Attn}q) \\ 
    \widehat{f} &= \sum \alpha_i f_i. 
    \end{split}
\end{equation}
$W_{Attn}$ represents the fully connected layer of the attention mechanism. $\alpha_i$ is the weight for the $i$th feature for fusing.

\noindent\textbf{Vision Embedding} 
As mentioned above, the panoramic observation $O_t$ denotes the 36 features consisting of vision and orientation information. 
We then fuse $\{o_{t,1}, ... ,  o_{t,36}\}$ with cross-modal context of the last step $\widehat{f}_{t-1}$ and introduce an LSTM to maintain a vision history context $\widetilde{f}^{o}_{t}$ for each step: 
\begin{equation}
\begin{split}
\widehat{f}^o_{t} &= \textnormal{Attn}_o(\{o_{t,1}, ... ,  o_{t,36}\} , \widetilde{f}_{t-1}) \\
\widetilde{f}_{t}^{o} &= \textnormal{LSTM}_v(\widehat{f}^o_{t}, h_{t-1}), 
\end{split}
\label{eq:visoin-embedding}
\end{equation}
where $\widetilde{f}_{t}^{o}=h_t$ is the output of the $\textnormal{LSTM}_v$. Note that unlike the other two LSTM layers in our pipeline (as shown in Fig.~\ref{fig:pipeline}) which are computed within a step.  $\textnormal{LSTM}_v$ is computed over a whole trajectory. 

\noindent\textbf{Vision-Language Embedding}
Similar to~\cite{fried2018speaker, tan2019learning}, we embed each word token $w_i$ to word feature $f^{w}_{i}$, where $i$ stands for the index. Then we encode the feature sequence by a Bi-LSTM layer to produce language features and a global language context $\overline{f}^{w}$: 
\begin{equation}
\begin{split}
\{\widetilde{f}^{w}_{0} ,..., \widetilde{f}^{w}_{l}\} &= \textnormal{Bi-LSTM}_w(\{f^{w}_{0} ,..., f^{w}_{l}\}) \\
\overline{f}^{w} &= \frac{1}{l} \sum_{i=1}^{l} \widetilde{f}^{w}_{i}. 
\end{split}
\label{eq:vl-embedding}
\end{equation}
The  global language context participates $\overline{f}^{w}$ the auxiliary task learning descripted in Sec.~\ref{Auxiliary Reasoning Learning}. Finally, we fuse the language features $\{\widetilde{f}^{w}_{0} ,..., \widetilde{f}^{w}_{l}\}$  with the vision history context $\widetilde{f}^o_t$ to produce the cross-modal context $\widehat{f}_{t}$: 
\begin{equation}
\label{eq:vle}
\widehat{f}_{t} = \textnormal{Attn}_w(\{\widetilde{f}^{w}_{0} ,..., \widetilde{f}^{w}_{l}\}, \widetilde{f}^{o}_{t}). 
\end{equation}

\noindent\textbf{Action Prediction}
In the VLN setting, the adjacent navigable node is visible. Thus, we can obtain the reachable candidates $C=\{c_{t,1},...,c_{t,k+1}\}$ from the navigation graph. Similar to observation $O$,  candidates in $C$ are concatenated features of vision features and orientation descriptions. 
We obtain the probability function $p_t(a_t)$ for action $a_t$ by: 
\begin{equation}
\begin{split}
\widehat{f}^{c}_{t} &= \textnormal{Attn}_c(\{c_{t,1},...,c_{t,k+1}\}, \widehat{f}_{t}) \\
p_t(a_t) &= \textnormal{softmax}(\widehat{f}^{c}_{t}). 
\end{split}
\end{equation}
Three ways for action prediction are applied to different scenarios: 1) imitation learning: following the labeled teacher action $a_t^*$ regardless of $p_t$; 2) reinforcement learning: sample action following the probability distribution $a_t \sim p_t(a_t)$; 3) testing: choose the candidate which has the greatest probability $a_t = \textnormal{argmax}(p_t(a_t))$. 

\subsection{Objectives for Navigation}\label{Imitation and Reinforcement Learning}
In this section, we introduce two learning objectives for the navigation task: imitation learning (IL) and reinforcement learning (RL). The navigation task is jointly optimized by these two objectives. 

\noindent\textbf{Imitation Learning} forces the agent to mimic the behavior of its teacher. IL has been proven~\cite{fried2018speaker} to achieve good performance in VLN tasks. 
Our agent learns from the teacher action $a_t^*$ for each step: 
\begin{equation}
L_{IL} = \sum_t - a_t^* \textnormal{log}(p_t), 
\end{equation}
where $a_t^*$ is a one-hot vector indicating the teacher choice. 

\noindent\textbf{Reinforcement Learning} is introduced for generalization since adopting IL alone could result in overfitting. 
We implement the A2C algorithm, the parallel version of A3C~\cite{mnih2016asynchronous}, and our loss function is calculated as: 
\begin{equation}
L_{RL} = - \sum_t a_t \textnormal{log}(p_t)  A_t. 
\end{equation}
$A_t$ is a scalar representing the advantage defined in A3C. 

\noindent\textbf{Joint Optimization} Firstly, the model samples trajectory
by teacher forcing approach and calculates gradients with imitation learning. Secondly, the model samples trajectory under the same instruction by student forcing approach and calculates gradients with reinforcement learning. Finally,
we add the gradients together and use the added gradients to update the model. 

\subsection{Auxiliary Reasoning Learning}\label{Auxiliary Reasoning Learning}
The vision-language navigation task remains challenging, since the rich semantics contained in the environments are neglected. 
In this section, we introduce auxiliary reasoning learning to exploit additional training signals from environments. 

In Sec.~\ref{Vision-Language Forward}, we obtain the vision context $\widetilde{f}^{o}_{t}$ from Eq.~\ref{eq:visoin-embedding}, the global language context $\overline{f}^{w}$ from Eq.~\ref{eq:vl-embedding} and the cross-modal context $\widehat{f}_{t}$ from Eq.~\ref{eq:vle}. In addition to action prediction, we give the contexts to the reasoning modules in Fig.~\ref{fig:pipeline} to perform auxiliary tasks. 
We discuss four auxiliary objectives use the contexts for reasoning below. 

\noindent\textbf{Trajectory Retelling Task} 
Trajectory reasoning is critical for an agent to decide what to do next. 
Previous works train a speaker to translate a trajectory to a language instruction. The methods are not end-to-end optimized, which limit the performances. 

As shown in Fig.~\ref{fig:pipeline}, we adopt a teacher forcing method to train an end-to-end speaker. The teacher is defined as $\{f^{w}_{0} ,..., f^{w}_{l}\}$, the same word embeddings as in Eq.~\ref{eq:vle}. We use $\textnormal{LSTM}_s$ to encode these word embeddings. We then introduce a cycle reconstruction objective named trajectory retelling task: 
\begin{equation}
\begin{split}
   \{\widetilde{f}^{w}_{0} ,..., \widetilde{f}^{w}_{l}\} &= \textnormal{LSTM}_s(\{f^{w}_{0} ,..., f^{w}_{l}\}), \\
   \widehat{f}^{s}_{i} &= \textnormal{Attn}_s(\{\widetilde{f}^{o}_{0},..., \widetilde{f}^{o}_{T}\}, \widetilde{f}^{w}_{i}), \\ 
   L_{Speaker} &= - \frac{1}{l} \sum_{i=1}^{l} \textnormal{log } p(w_i |\widehat{f}^{s}_{i}). 
\end{split}
\end{equation}
Our trajectory retelling objective is jointly optimized with the main task. It helps the agent to obtain better feature representations since the agent comes to know the semantic meanings of the actions. Moreover, trajectory retelling makes the activity of the agent explainable. Since the model could deviate a lot in student forcing, we does not train the trajectory retelling task in RL scenarios. 

\noindent\textbf{Progress Estimation Task}
We propose a progress estimation task to learn the navigation progress. Earlier research~\cite{ma2019self} uses normalized distances as labels and optimizes the prediction module with Mean Square Error (MSE) loss. However, we use the percentage of steps $r_t$, noted as a soft label $\{\frac{t}{T}, 1-\frac{t}{T}\}$ to represent the progress: 
\begin{equation}
   L_{progress} = - \frac{1}{T} \sum_{t=1}^{T} r_t \textnormal{log } \sigma( W_r \widehat{f}_{t}).
\end{equation}
Here $W_r$ is the weight of the fully connected layer and $\sigma$ is the sigmoid activation layer. 
Our ablation study reveals that the method that learning from percentage of steps $r_t$ with BCE loss achieves higher performance than previous method. 
Normalized distance labels introduce noise, which limits performance. Moreover, we also find that Binary Cross Entropy (BCE) loss performs better than MSE loss with  our step-percentage label since logits learned from BCE loss are unbiased. The progress estimation task requires the agent to align the current view with corresponding words in the instruction. Thus, it is beneficial to vision language grounding. 

\begin{table*}[t]
\small
 \begin{center}
 \resizebox{1.0\textwidth}{!}{
 \setlength{\tabcolsep}{1.0em}
 {\renewcommand{\arraystretch}{1.0}
  \begin{tabular}{|l | c c  c c | c c c c | c  c c | }
   \hline
    Leader-Board (Test Unseen) 
    & \multicolumn{4}{c|}{Single Run} 
    & \multicolumn{4}{c|}{Pre-explore}
    & \multicolumn{3}{c|}{Beam Search}\\
   \hline
   Models
 & NE & OR & SR & \emph{SPL}
 & NE & OR & SR & \emph{SPL}
 & TL & SR & \emph{SPL}\\
 \hline
 Random~\cite{anderson2018vision} &  9.79 &  0.18 &  0.17 & 0.12  & - & - & - & - & - & - & - \\
 Seq-to-Seq~\cite{anderson2018vision} &  20.4 &  0.27 &  0.20 & 0.18  & - & - & - & - & - & - & -\\
 \hline
 Look Before You Leap~\cite{wang2018look} &  7.5 &   0.32 & 0.25  & 0.23  &-   &-  &- &- & - & - & -\\
 Speaker-Follower~\cite{fried2018speaker} &  6.62 &   0.44 & 0.35  & 0.28  &-  &-  &- &- &1257 &0.54 &0.01\\
 Self-Monitoring~\cite{ma2019self} & 5.67 & 0.59 & 0.48  & 0.35  &-  &-  &- &- &373 &0.61 &0.02\\
 The Regretful Agent~\cite{ma2019regretful} &  5.69 &   0.48 & 0.56  & 0.40  &-  &-  &- &- &13.69 &0.48 &0.40\\
 FAST~\cite{ke2019tactical} &  5.14 &   - & 0.54  & 0.41  &-  &-  &- &- & 196.53 & 0.61 &0.03\\
 Reinforced Cross-Modal~\cite{wang2018reinforced} & 6.12 &  0.50 & 0.43  & 0.38 & 4.21 &  0.67 & 0.61  & 0.59 &358 &0.63 &0.02 \\
 ALTR~\cite{huang2019transferable} & 5.49 &  - & 0.48  & 0.45 & - &  - & -  & - &- &- &-\\
 Environmental Dropout~\cite{tan2019learning} & 5.23 &  0.59 & 0.51  & 0.47 & 3.97 &  0.70 & 0.64  & 0.61 &687 &0.69 &0.01\\
 \hline
 AuxRN(Ours) & \textbf{5.15} &  \textbf{0.62} & \textbf{0.55}  & \textbf{0.51} & \textbf{3.69} &  \textbf{0.75} & \textbf{0.68}  & \textbf{0.65} & 41 
 &\textbf{0.71} &\textbf{0.21} \\
   \hline
 \end{tabular}
 }
 }
 \end{center}
 \caption{
 Leaderboard results comparing AuxRN with the previous state-of-the-art on test split in unseen environments. We compare three training settings:  Single  Run  (without seeing unseen environments),  Pre-explore  (finetuning in unseen environments), and  Beam  Search(comparing success rate regardless of TL and SPL). The primary metric for Single Run and Pre-explore is SPL, while the primary metric for Beam Search is the success rate (SR).  We only report two decimals due to the precision limit of the leaderboard. 
 }
 \vspace{-9pt}
 \label{table:result}
 \end{table*}

\noindent\textbf{Cross-modal Matching Task}
We propose a binary classification task, motivated by LXMERT~\cite{tan2019lxmert}, to predict whether or not the trajectory matches the instruction. 
We shuffle $\overline{f}^{w}$ from Eq.~\ref{eq:vl-embedding} with  feature vector in the same batch with the probability of $0.5$.
The shuffled operation is marked as ``S" in the white circle in Fig.~\ref{fig:pipeline} and the shuffled feature is noted as $\overline{f{}'}^{w}$. 
We concatenate the shuffled feature with the attended vision-language feature $\widehat{f}_{t}$. We then supervise the prediction result with $m_t$, a binary label indicating whether the feature has been shuffled or remains unchanged. 
\begin{equation}
   L_{Matching} = - \frac{1}{T} \sum_{t=1}^{T} m_t \textnormal{log } \sigma( W_m [\widehat{f}_{t}, \overline{f{}'}^{w}]), 
\end{equation}
where $W_m$ stands for the fully connected layer. This task requires the agent to align historical vision-language features in order to distinguish if the overall trajectory matches the instruction. Therefore, it facilitates the agent to encode historical vision and language features. 

\noindent\textbf{Angle Prediction Task}
The agent make the choice among the candidates to decide which step it will take next. 
Compared with the noisy vision feature, the orientation is much cleaner.
Thus we consider learning from orientation information in addition to learning from candidate classification. 
We thus propose a simple regression task to predict the orientation that the agent will turn to: 
\begin{equation}
   L_{angle} = - \frac{1}{T} \sum_{t=1}^{T} 
   \parallel e_t - W_e\widehat{f}_{t} \parallel, 
\end{equation}
where $a_t$ is the angle of the teacher action in the imitation learning, while $W_a$ stands for the fully connected layer. Since this objective requires a teacher angle for supervision, we do not forward this objective in RL. 

Above all, we jointly train all the four auxiliary reasoning tasks in an end-to-end manner: 
\begin{equation}
   L_{total} =  L_{Speaker}+L_{Progress}+L_{Angle}+L_{Matching}.
\end{equation}
\section{Experiment}
\subsection{Setup}
\noindent\textbf{Dataset and Environments} We evaluate the proposed AuxRN method on the Room-to-Room (R2R) dataset~\cite{anderson2018vision} based on Matterport3D simulator~\cite{chang2017matterport3d}. The dataset, comprising 90 different housing environments, is split into a training set, a seen validation set, an unseen validation set and a test set.  
The training set consists of 61 environments and 14,025 instructions, while the seen validation set has 1,020 instructions using the save environments with the training set. The unseen validation set consists of another 11 environments with 2,349 instructions, while the test set consists of the remaining 18 environments with 4,173 instructions. 

\noindent\textbf{Evaluation Metrics} 
A large number of metrics are used to evaluate models in VLN, such as Trajectory Length (TL), the trajectory length in meters, Navigation Error (NE), the navigation error in meters, Oracle Success Rate (OR), the rate if the agent successfully stops at the closest point, Success Rate (SR), the success rate of reaching the goal, and Success rate weighted by (normalized inverse) Path Length (SPL)~\cite{anderson2018on}. In our experiment, we take all of these into consideration and regard SPL as the primary metric. 

\noindent\textbf{Implementation Details} 
We introduce self-supervised data to augment our dataset.  We sample the augmented data from training and testing environments and use the speaker trained in Sec.~\ref{Vision-Language Forward} to generate self-supervised instructions. 

Our training process consists of three steps: 1) we pretrain our model on the training set; 2) we pick the best model (the model with the highest SPL) at step 1 and finetune the model on the augmented data sampled from training set~\cite{tan2019learning}; 3) we finetune the best model at step 2 on the augmented data sampled from testing environments for pre-exploration, which is similar to~\cite{wang2018reinforced, tan2019learning}. We pick the last model at step 3 to test. 
The training iterations for each steps are 80K. We train each model with auxiliary tasks and set all auxiliary loss weight to 1. 
At steps 2 and 3, since augmented data contains more noise than labeled training data, we reduce the  loss weights for all auxiliary tasks by half. 

\begin{table*}[t]
\small
 \begin{center}
 \resizebox{1.0\textwidth}{!}{
 \setlength{\tabcolsep}{1.3em}
 {\renewcommand{\arraystretch}{1.0}
  \begin{tabular}{|l | c c  c c | c c  c c | }
  \hline
     &  \multicolumn{4}{c|}{Val Seen} 
        & \multicolumn{4}{c|}{Val Unseen}\\
  \hline
  Models
  & NE (m) & OR (\%) & SR (\%) & \emph{SPL} (\%)
  & NE (m) & OR (\%) & SR (\%) & \emph{SPL} (\%)\\
 \hline
  baseline & 4.51 &  65.62 &  58.57 & 55.87 &  5.77 & 53.47 &  46.40 & 42.89 \\
 \hline
 baseline+$L_{Speaker}$ &  4.13 &  69.05 &  60.92 & 57.71 &  5.64 &  57.05 &  49.34 & 45.24 \\
 baseline+$L_{progress}$ &  4.35 &  68.27 & 60.43 & 57.15 & 5.80 & 56.75 & 48.57 & 44.74 \\
 baseline+$L_{Matching}$ &  4.70 &  65.33 & 56.51 & 53.55 & 5.74 & 55.85 & 47.98 & 44.10 \\
 baseline+$L_{Angle}$ &  4.25 &  70.03 & 60.63 & 57.68 & 5.87 & 55.00 & 47.94 & 43.77 \\
 baseline+$L_{Total}$ &  4.22 &  72.28 & 62.88 & \textbf{58.89}  & 5.63 & 59.60 & 50.62 & \textbf{45.67} \\
 \hline
 baseline+BT~\cite{tan2019learning}  &  4.04 &  70.13 & 63.96 & 61.37 & 5.39 & 56.62 & 50.28 & 46.84 \\
 baseline+BT+$L_{Total}$  &  \textbf{3.33} &  \textbf{77.77} & \textbf{70.23} & \textbf{67.17} & \textbf{5.28} & \textbf{62.32} & \textbf{54.83} & \textbf{50.29} \\
 \hline
 \end{tabular}
 }
 }
 \end{center}
 \caption{
 Ablation study for different auxiliary reasoning tasks. We evaluate our models on two validation splits: validation for the seen and unseen environments. 
 Four metrics are compared, including NE, OR, SR and SPL. 
 }
 \vspace{-9pt}
 \label{table:abla_all}
 \end{table*}

\subsection{Test Set Results}

In this section, we compare our model with previous state-of-the-art methods.  We compare the proposed AuxRN with two baselines and five other methods. A brief description of previous models as followed. 1) Random: randomly take actions for 5 steps. 2) Seq-to-Seq: A sequence to sequence model reported in~\cite{anderson2018vision}. 3) Look Before You Leap: a method combining model-free and model-based reinforcement learning. 4) Speaker-Follower: a method introduces a data augmentation approach and panoramic action space. 5) Self-Monitoring: a method regularized by a self-monitoring agent. 6) The Regretful Agent: a method based on learnable heuristic search 7) FAST: a search based method enables backtracking 8) Reinforced Cross-Modal: a method with cross-modal attention and combining imitation learning with reinforcement learning.  9) ALTR: a method focus on adapting vision and language representations 10) Environmental Dropout: a method augment data with environmental dropout. Additionally, we evaluate our models on three different training settings: 1) Single Run: without seeing the unseen environments and 2) Pre-explore: finetuning a model in the unseen environments with self-supervised approach. 3) Beam Search: predicting the trajectories with the highest rate to success.

As shown in Tab.~\ref{table:result}, AuxRN outperforms previous models in a large margin on all three settings. 
In Single Run, we achieve 3\% improvement on oracle success, 4\% improvement on success rate and 4\% improvement on SPL. 
In Pre-explore setting, our model greatly reduces the error to 3.69, which shows that AuxRN navigates further toward the goal. AuxRN significantly boost oracle success by 5\%, success rate 4\% and SPL to 4\%. AuxRN achieves similiar improvements on other two domains, which indicates that the auxiliary reasoning tasks is immune from domain gap. 

We also achieve the state-of-the-art in Beam Search setup. Our final model with Beam Search algorithm achieves 71\% success rate, which is 2\% higher than Environmental Dropout, the previous state-of-the-art. 

\subsection{Ablation Experiment}

\noindent\textbf{Auxiliary Reasoning Tasks Comparison} In this section, we compare  performances between different auxiliary reasoning tasks. 
We use the previous state-of-the-art~\cite{tan2019learning} as our baseline. We train the models with each single task based on our baseline. 
We evaluate our models on both the seen and unseen validation set and the results are shown in Tab.~\ref{table:abla_all}. 
It turns out that each task promotes the performance based on our baseline independently. And training all tasks together is able to further boost the performance, achieving improvements by 3.02\% on the seen validation set and by 2.78\% on the unseen validation set. It indicates that the auxiliary reasoning tasks are presumably reciprocal. 

Moreover, our experiments show that our auxiliary losses and back-translation method has a mutual promotion effect. On the seen validation set, baseline with back-tranlation gets 5.50\% improvement while combining back-translation promotes SPL by 11.30\%, greater than the sum of the performance improvement of baseline with auxiliary losses and with back-translation independently. Similar results have been observed on the unseen validation set. Baseline with back-translation gets 3.95\% promotion while combining back-translation boosts SPL by 7.40\%. 

\begin{table}[t]
\small
 \begin{center}
\resizebox{1.0\linewidth}{!}{
\setlength{\tabcolsep}{0.6em}
{\renewcommand{\arraystretch}{1.0}
\begin{tabular}{|l|l|c c c c|}
\hline
& Models & OR(\%) & SR(\%)  & Acc(\%) & SPL(\%) \\ \hline
\multirow{5}{*}{\rotatebox[origin=c]{90}{Val Seen}}  &  Baseline & 65.62   & 58.57   & -  & 55.87        \\ \cline{2-6} 
&  Matching Critic~\cite{wang2018reinforced} & 63.76  & 55.73   & 19.58 & 52.77          \\ 
& Student Forcing~\cite{anderson2018vision} & 65.72  & 57.59  & 25.37 & 54.95         \\ 
& Teacher Forcing(share)   & 66.90  & 60.33   & \textbf{34.85} & \textbf{57.23}          \\ 
&  Teacher Forcing(ours) & 65.62  & 59.55 & 26.34 & 56.99           \\ 

\hline
\multirow{5}{*}{\rotatebox[origin=c]{90}{Val Unseen}} &  Baseline & 53.47   & 46.40   & -  & 42.89        \\ \cline{2-6} 
&  Matching Critic & 55.26  & 46.74  & 18.88 & 43.44          \\ 
& Student Forcing & 54.92  & 47.42  & 25.04 & 43.78          \\ 
& Teacher Forcing(share)   & 56.41  & 48.19  & \textbf{38.49} & 44.47          \\ 
&  Teacher Forcing & 57.05  & 49.34   & 25.95  & \textbf{45.24}        \\ 
\hline
\end{tabular}
}
}
\end{center}
\caption{
 Ablation study for Trajectory Retelling Task. 
 Four metrics are compared, including OR, SR, SPL and Acc (sentence prediction accuracy). 
 }
 \vspace{-13pt}
 \label{table:abla_speaker}
\end{table}

\noindent\textbf{Ablation for Trajectory Retelling Task}
We evaluate four different implementations for trajectory retelling task. All method uses visual contexts for trajectories to predict word tokens. 1) Teacher Forcing: The standard Trajectory Retelling approach as described in Sec.~\ref{Auxiliary Reasoning Learning}. 2) Teacher Forcing(share): an variant of teacher forcing which uses $\Tilde{f}_w$ to attend visual features. 3) Matching Critic: regards opposite number of the speaker loss as a reward to encourage the agent. 4) Student Forcing: a seq-to-seq approach translating visual contexts to word tokens without ground truth sentence input. In addition to OR, SR, and SPL, we add a new metric, named sentence prediction accuracy (Acc). This metric calculates the precision model predict the correct word. 

The result of ablation study for Trajectory Retelling Task is shown as Tab.~\ref{table:abla_speaker}. 
Firstly, teacher forcing outperforms Matching Critic~\cite{wang2018reinforced} by 1.8\% and 4.22\% respectively. Teacher forcing performs 7.07\% and 6.76\% more than  Matching Critic in terms of accuracy. 
Secondly, teacher forcing outperforms student forcing by 1.46\% and 2.04\% in terms of SPL in two validation sets. The results also indicate that teacher forcing is better in sentence prediction compared with student forcing. 
Thirdly, in terms of SPL, standard teacher forcing outperforms the teacher forcing with shared context on the unseen validation set by 0.77\%. Besides, we notice that the teacher forcing with shared context outperforms standard teacher forcing about 12\% in word prediction accuracy (Acc). We infer that the teacher forcing with shared context overfits on the trajectory retelling task.

\begin{table}[t]
\small
 \begin{center}
\resizebox{1.0\linewidth}{!}{
{\renewcommand{\arraystretch}{1.0}
\begin{tabular}{|l|l|c c c c|}
\hline
& Models & OR(\%) & SR(\%)  & Error & SPL(\%) \\ \hline
\multirow{4}{*}{\rotatebox[origin=c]{90}{Val Seen}}  &  Baseline & 65.62   & 58.57   & -  & 55.87        \\ \cline{2-6} 
&  Progress Monitor~\cite{ma2019self} & 66.01  & 57.1   & 0.72 & 53.43          \\ 
& Step-wise+MSE(ours)  & 64.15  & 53.97   & 0.27 & 50.81         \\ 
& Step-wise+BCE(ours) & \textbf{68.27}  & \textbf{60.43}   & \textbf{0.13} & \textbf{57.15}           \\ 

\hline
\multirow{4}{*}{\rotatebox[origin=c]{90}{Val Unseen}} &  Baseline & 53.47   & 46.40   & -  & 42.89        \\ \cline{2-6} 
&  Progress Monitor & \textbf{57.09}  & 46.57  & 0.80 & 42.21          \\ 
& Step-wise+MSE(ours)   & 55.90  & 46.74  & 0.32 & 43.16          \\ 
& Step-wise+BCE(ours) & 56.75  & \textbf{48.57}  & \textbf{0.16} & \textbf{44.74}        \\ 
\hline
\end{tabular}
}
}
\end{center}
\caption{
 Ablation study for Progress Estimation Task. 
 Four metrics are compared, including OR, SR, SPL and Error (normalized absolute error). 
 }
 \vspace{-13pt}
 \label{table:abla_progress}
\end{table}

\noindent\textbf{Progress Estimation Task} 
To valid the progress estimation task, we investigation two variants in addition to our standard progress estimator. 1) Progress Monitor: We implement Progress Monitor~\cite{ma2019self} based on our baseline method.  2)  we train our model use Mean Square Error (MSE) rather than BCE Loss with the same step-wise label $\frac{t}{T}$. We compare these models with four metris: OR, SR, Error and SPL. The Error is calculated by the mean absolute error between the progress estimation prediction and the label. 

The result is shown as Tab.~\ref{table:abla_progress}. Our standard model outperforms other two variants and the baseline on most of the metrics. Our Step-wise MSE model performs 2.62\% higher on the seen validation set 2.53\% higher on the unseen validation set than Progress Monitor~\cite{ma2019self}, indicating that label measured by normalized distances is noisier than label measured by steps. In addition, we find that the Progress Monitor we implement performs even worse than baseline. 
When the agent begins to deviate from the labeled path, the progress label become even noisier.  

We compare different loss functions with step-wise labels. Our model with BCE loss is 6.34\% higher on the seen validation set and 1.58\% higher on the unseen validation set. Furthermore, the prediction error of the model trained by MSE loss is higher than which trained by BCE loss. The Error of the Step-wise+MSE model is 0.14 higher on the seen validation set and 0.16 higher on the unseen validation set than Step-wise+BCE model. 

\begin{figure}[t]
	\centering
	\includegraphics[width=0.95\linewidth]{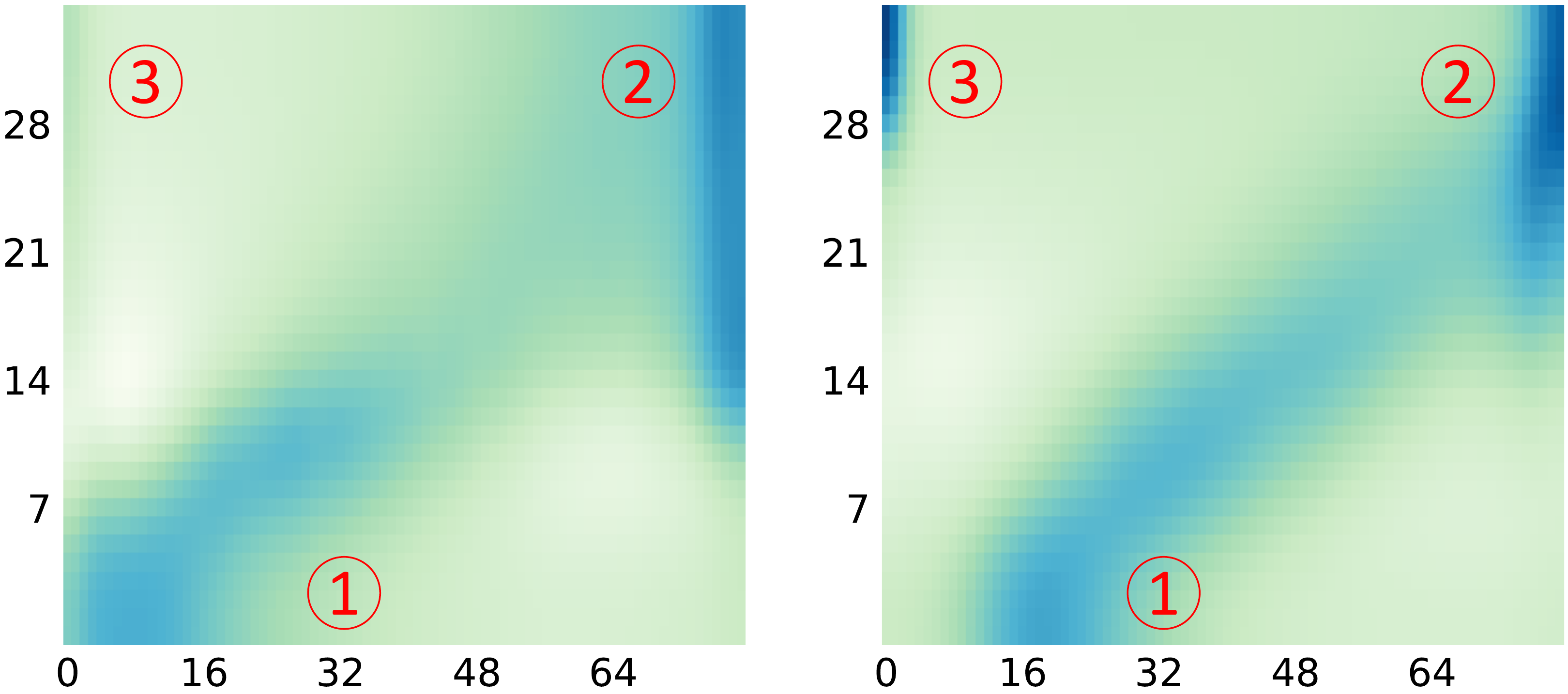}
	\caption{The language attention map for the baseline model and our final model. The x-axis stands for the position of words and the y-axis stands for the navigation time steps. Since each trajectory has variable number of words and number of steps, we normalize each attention map to the same size before we sum all the maps. 
	}
	\label{fig:heatmap}
	\vspace{-0.5cm}
\end{figure}

\begin{figure*}[t]
	\centering
	\includegraphics[width=0.90\linewidth]{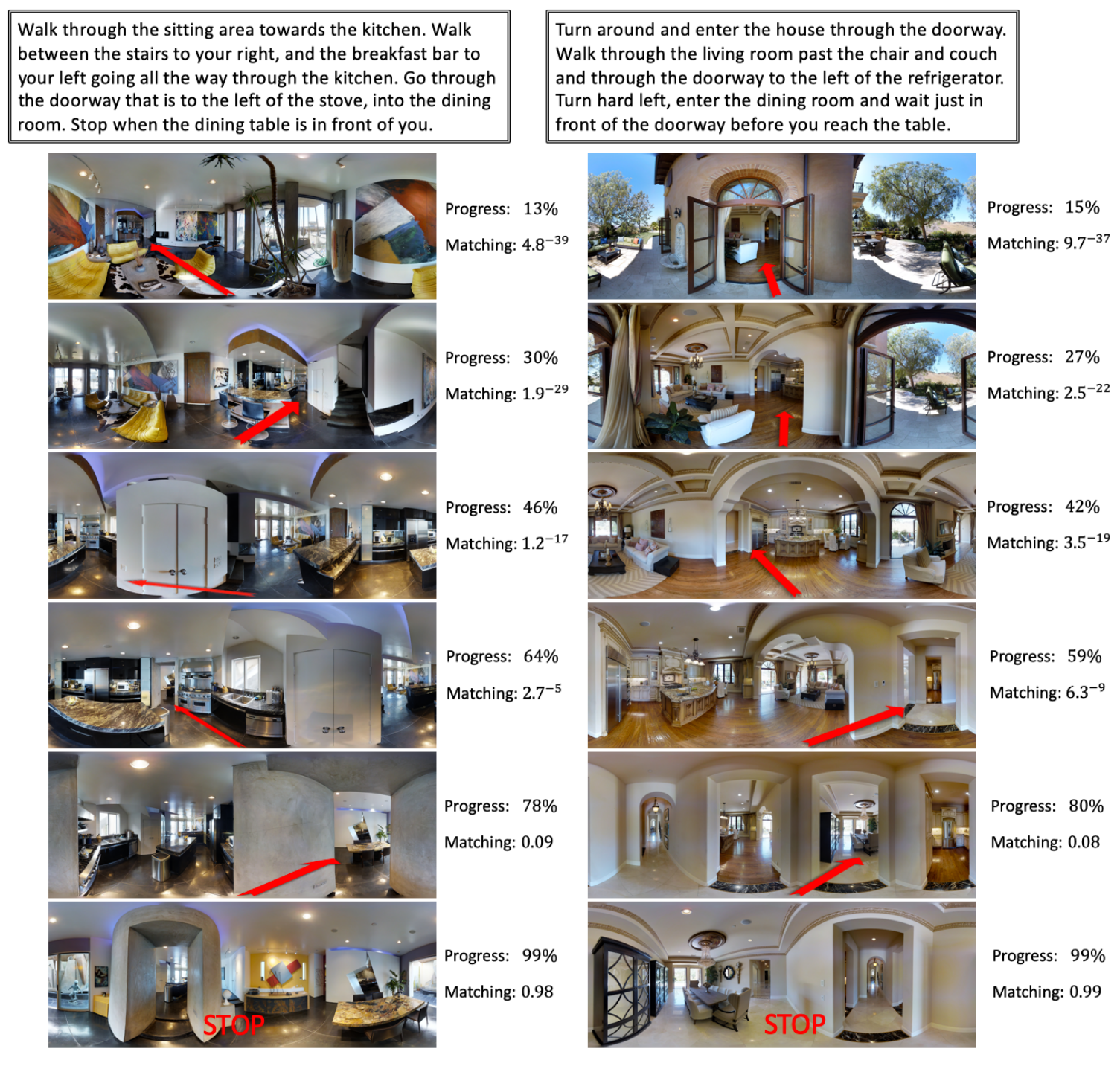}
	\caption{
	Visualization process of two trajectories in testing. Two complex language instructions are shown in top boxes. Each image is a panoramic view, which is the vision input for AuxRN. Each red arrow represents the direction to the next step. 
	For each step, the results progress estimator and the matching function are shown as left. 
	}
	\label{fig:attention}
	\vspace{-0.5cm}
\end{figure*}

\subsection{Visualization}
\noindent\textbf{Regularized Language Attention}~\label{Regularized Language Attention}
We visualize the attention map for $\textnormal{Attn}_w$ after $\textnormal{Bi-LSTM}_w$. 
The dark region in the map stands for where the language features receive high attention. We observe from Fig.~\ref{fig:attention} that the attention regions on both two maps go left while the navigation step is increasing (marked as 1). It means that both models learns to pay an increasing attention to the latter words. At the last few steps,  our model learns to focus on the first feature and the last feature (marked as 2 and 3), since the Bi-LSTM encodes sentence information at the first and the last feature. 
We infer from our experiments that auxiliary reasoning losses help regularize the language attention map, which turns out to be beneficial. 


\noindent\textbf{Navigation Visualization} We visualize two sample trajectories to show the process of navigation. To further demonstrate how AuxRN understand the environment, we show the result of the progress estimator and matching function. The estimated progress continues growing during navigation while the matching result is increasing exponentially. When AuxRN reaches the goal, the progress and matching results jump to almost 1. It turns out that our agent precisely estimating the current progress and the instruction trajectory consistency. 

\section{Conclusion}

In this paper, we presented a novel framework, auxiliary Reasoning Navigation (AuxRN), that facilitates navigation learning with four auxiliary reasoning tasks. 
Our experiments confirm that AuxRN improves the performance of the VLN task quantitatively and qualitatively. We plan to build a general framework for auxiliary reasoning tasks to exploit the common sense information in the future.

\section*{Acknowledgement}

This work was supported in part by the National Natural Science Foundation of China (NSFC) under Grant
No.U19A2073 and in part by the National Natural Science Foundation of China (NSFC) under Grant No.61976233, and by the Air Force Research Laboratory and DARPA under agreement number FA8750-19-2-0501, Australian Research Council Discovery Early Career Researcher Award (DE190100626).

\clearpage


\end{document}